\documentclass[letterpaper,10pt,conference]{ieeeconf}
\IEEEoverridecommandlockouts
\usepackage{pifont}
\usepackage{graphics} 
\usepackage{graphicx,tabularx,adjustbox}
\usepackage{tikz}
\usetikzlibrary{fadings}
\usetikzlibrary{calc,arrows.meta,decorations.pathreplacing,positioning}
\usepackage{multirow}
\usepackage{booktabs}
\usepackage{algorithm}
\usepackage{algpseudocode}
\algnewcommand{\LineComment}[1]{\Statex \(\triangleright\)\ #1}
\usepackage{hhline}
\usepackage{xcolor}
\usepackage{colortbl}
\usepackage{float}
\usepackage{subcaption}
\usepackage{amsmath} 
\usepackage{amssymb}  
\usepackage{amsfonts,amstext,dsfont,mathtools,bbm}
 
\usepackage{amsthm}
\usepackage{balance}
\usepackage{cite}
\usepackage{xspace}
\usepackage{pgfplots}
\pgfplotsset{compat=1.18}

\makeatletter\let\NAT@parse\undefined\makeatother
\usepackage[bookmarks=true, colorlinks, breaklinks=true]{hyperref}
\usepackage{xurl} 

\definecolor{oursrow}{gray}{0.88}

\definecolor{oursrow}{RGB}{220,239,245}      
\definecolor{groupgray}{RGB}{245,245,245}    
\definecolor{headerblue}{RGB}{223,235,242}   

\newcommand{\method}{AdaGeoVLN}

\newcommand{\cmark}{\ding{51}}

\title{\LARGE \bf
AdaGeoVLN: Selective Geometry Across Representation Depth and Navigation Time for Vision-Language Navigation
}
\author{Quan-Dung Pham$^{1, *}$, Anh Dao$^{1, *}$, Danh Vinh Le$^{1, *}$, Nguyen Viet Tri Pham$^{1, *}$, The-Anh Nguyen$^{1}$, \\ Zhirui Dai$^{2}$, Yiyu Chen$^{1}$, Tuyen P. Le$^{1}$, Truong Nguyen$^{1}$, Quan Nguyen$^{3}$
\thanks{$^{1}$VinMotion, Inc., Vietnam}
\thanks{$^{2}$Department of Electrical and Computer Engineering, University of California San Diego, La Jolla, CA 92093, USA}
\thanks{$^{3}$University of Southern California, USA}
\thanks{$^{*}$Equal contribution}
}

\begin{document}
\IEEEaftertitletext{%
    \vspace{2pt}%
    \begin{center}
        \includegraphics[width=0.95\textwidth]{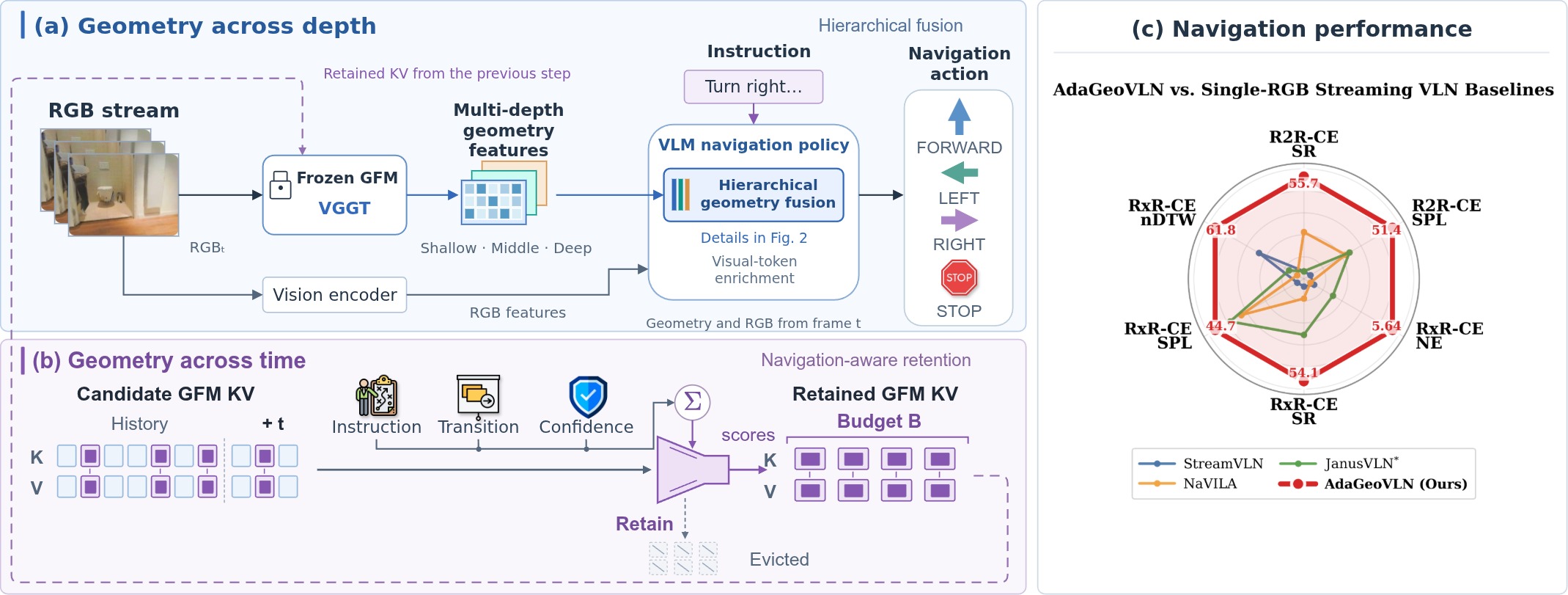}
        \captionof{figure}{\textbf{AdaGeoVLN overview.} Geometry is selected along two complementary axes. Across \emph{representation depth}, intermediate VGGT states are coupled to successive VLM stages. Across \emph{navigation time}, historical VGGT global-attention KV states are selectively retained under a bounded memory budget. The right panel summarizes performance relative to representative single-RGB streaming VLN baselines on R2R-CE and RxR-CE Val-Unseen.}
        \label{fig:overview}
    \end{center}
    \vspace{2pt}%
}
\maketitle
\thispagestyle{empty}
\pagestyle{empty}

\begin{abstract}
Vision-language navigation requires aligning language with visual observations while maintaining spatial understanding over time. Geometry foundation models (GFMs) expose intermediate representations throughout their hierarchy, but how navigation policies should use these features and retain historical geometric evidence remains unresolved. We introduce \method{}, a streaming VLN framework that addresses these questions across \textbf{representation depth} and \textbf{navigation time}. Hierarchical GFM--VLM fusion couples earlier, intermediate, and later GFM representations to successive policy stages instead of repeatedly injecting a terminal feature. Navigation-aware GFM memory retains historical VGGT global-attention KV states according to instruction relevance, geometric confidence, and transition novelty under a bounded per-layer budget. Retained states provide geometric context for subsequent observations before fusion with the policy. Across R2R-CE and RxR-CE, \method{} achieves strong performance using a single RGB stream without additional navigation-specific external data. Controlled ablations show that multi-depth coupling substantially outperforms repeated terminal-feature injection at matched fusion locations. Bounded navigation-aware retention preserves navigation performance while considerably reducing GFM-KV memory relative to larger-memory temporal retention. These findings support jointly examining the geometric representations exposed to the policy and the historical evidence retained for future inference. \textit{Project page:} \url{https://humanoid-research.github.io/adageovln/}.
\end{abstract}

\section{INTRODUCTION}
\label{sec:introduction}

Vision-language navigation (VLN) has advanced through stronger vision-language models, large-scale instruction following, and streaming policies operating on continuous visual observations. Yet navigating unseen environments requires more than object recognition or language--appearance matching. An agent must reason about spatial relations, viewpoint changes, route geometry, and temporal connections among observations to follow an instruction. These demands make spatial understanding central to VLN and motivate geometric representations that complement visual-semantic reasoning.

Geometry foundation models (GFMs) provide a promising source of spatial information. VGGT~\cite{wang2025vggt}, for example, learns geometric representations from large-scale visual data and infers scene structure without requiring depth or pose inputs at deployment. However, \emph{how its representations should interact with a language-conditioned navigation policy remains unresolved}. A common approach exposes the policy only to the geometry encoder's terminal representation, although intermediate states may contain complementary information. This raises our first question: \emph{should representations from multiple geometric depths interact with successive stages of policy reasoning?} The distinction concerns representation diversity rather than injection frequency: repeatedly injecting a terminal feature increases interaction without exposing earlier GFM states. We therefore investigate \textbf{geometry selection across representation depth}, coupling representations from different geometric depths to successive policy stages without assigning predefined semantic roles to individual layers. A repeated-terminal control at matched fusion locations tests whether multi-depth access provides benefits beyond additional geometry--policy interactions. This comparison evaluates the utility of intermediate geometric representations for sequential navigation rather than assuming that the terminal feature is sufficient.

An orthogonal challenge arises across \textbf{navigation time}. Historical geometric states preserve spatial context across viewpoints, but unrestricted retention requires memory that grows with trajectory length. Recency-based retention bounds this growth while treating observation age as a proxy for utility, which may be inadequate for instruction-conditioned navigation. Earlier observations can remain useful because they capture an instruction landmark, provide reliable geometry, or represent a distinctive trajectory transition. This motivates our second question: \emph{which historical geometric states should remain available for future geometric inference under a bounded memory budget?}

We address these questions with \method{}, a streaming VLN framework combining multi-depth GFM--VLM coupling with navigation-aware retention of VGGT global-attention states (Fig.~\ref{fig:overview}). Retention uses instruction relevance, geometric confidence, and transition importance. Because retained states are reused within VGGT, selection shapes the context available to subsequent observations before policy fusion. The mechanisms govern which representations the policy accesses and which historical evidence supports their computation.

Our contributions are:
\begin{itemize}
    \item \textbf{Geometry selection across representation depth.} We couple multiple GFM depths to successive policy stages and use matched repeated-terminal controls to isolate the contribution of representation diversity.
    \item \textbf{Navigation-conditioned geometric memory.} We introduce bounded retention of historical GFM states based on instruction relevance, geometric confidence, and trajectory transitions, allowing navigation utility to guide the context available for future geometric inference.
    \item \textbf{Controlled evaluation in simulation and demonstration on a physical humanoid.} We evaluate \method{} on R2R-CE and RxR-CE, separately ablate selection across depth and time, and deploy the resulting policy on a Unitree G1 humanoid.
\end{itemize}
\section{RELATED WORK}
\label{sec:related}

\paragraph{Vision-Language Navigation.}
VLN-CE~\cite{krantz2020beyond} extended VLN from oracle navigation graphs to continuous environments, where agents execute low-level actions without known topology or perfect localization. Multimodal foundation models have since enabled video-centric navigation in NaVid and Uni-NaVid~\cite{zhang2024navid,zhang2024uni}, vision-language planning coupled with locomotion in NaVILA~\cite{cheng2024navila}, streaming context in StreamVLN~\cite{wei2025streamvln}, and separate spatial-geometric and visual-semantic implicit memories in JanusVLN~\cite{zeng2026janusvln}. Building on these advances, we investigate how a pretrained GFM's representations should support policy reasoning and persist across navigation time.

\paragraph{Spatial and Geometric Representations for VLN.}
Explicit spatial structure addresses limitations of appearance-only reasoning. Structured Scene Memory~\cite{wang2021ssm} organizes visual-geometric observations for long-range reasoning; GridMM~\cite{wang2023gridmm} aggregates instruction-relevant history in a growing top-down grid; Bird's-Eye-View Scene Graph~\cite{liu2023bsg} integrates local BEV representations with a global scene graph; and Volumetric Environment Representation~\cite{liu2024ver} lifts multi-view observations into structured 3D cells. These approaches provide policies with task-specific spatial structures. \method{} instead examines the hierarchy already learned by a pretrained GFM: whether its terminal state suffices for navigation, or whether intermediate geometric representations should remain accessible at successive policy stages.

\paragraph{Geometry Foundation Models and Geometry-Augmented MLLMs.}
Geometric representation learning has evolved from task-specific depth and reconstruction networks toward transferable models. Depth Anything~\cite{yang2024depthanything} scales monocular-depth pretraining; DUSt3R~\cite{wang2024dust3r} predicts multi-view point maps; MASt3R~\cite{leroy2024mast3r} extends this formulation to geometry-grounded matching; and VGGT~\cite{wang2025vggt} unifies camera estimation, depth, point maps, and tracking in a feed-forward Transformer. For multimodal reasoning, VG-LLM~\cite{zheng2025vgllm} introduces video geometry priors, while SpatialStack~\cite{zhang2026spatialstack} explores multi-level geometry. These studies motivate geometry-language coupling. We investigate its role in sequential embodied decision-making, specifically whether access to multiple GFM depths benefits VLN beyond repeatedly injecting a terminal representation.

\paragraph{Long-Horizon Memory and Streaming Navigation.}
VLN methods encode history through panoramic encoders~\cite{chen2021hamt}, explicit scene memories or maps~\cite{wang2021ssm,wang2023gridmm,chen2022duet}, and persistent environmental memory~\cite{krantz2023ivln}. Foundation-model policies also compress or reuse neural context: StreamVLN maintains compact streaming context~\cite{wei2025streamvln}, whereas JanusVLN retains fixed-size spatial-geometric and visual-semantic KV memories using initial and sliding windows~\cite{zeng2026janusvln}. \method{} selects the GFM's global-attention KV states according to instruction relevance, geometric confidence, and transition importance. Retention therefore determines which historical observations support \emph{future geometric inference}, allowing navigation utility to guide memory selection beyond recency alone.
\section{METHOD}
\label{sec:method}

\subsection{Problem Formulation and Overview}
At navigation step $t$, the agent observes an RGB image $I_t$ and instruction $\mathcal{L}$, then predicts $a_t\in\mathcal{A}$ from the current observation and history. A geometry foundation model (GFM) $\mathcal{G}$ processes the visual stream and exposes intermediate representations $\{G_t^m\}$ at depths $m$ to a vision-language navigation policy $\pi_\theta$:
\begin{equation}
 a_t \sim \pi_\theta(a_t\mid \mathcal{L}, I_{1:t}, \mathcal{M}_t),
\end{equation}
where $\mathcal{M}_t$ is the retained geometry-side memory.

\begin{figure}[!t]
    \centering
    \includegraphics[width=\columnwidth]{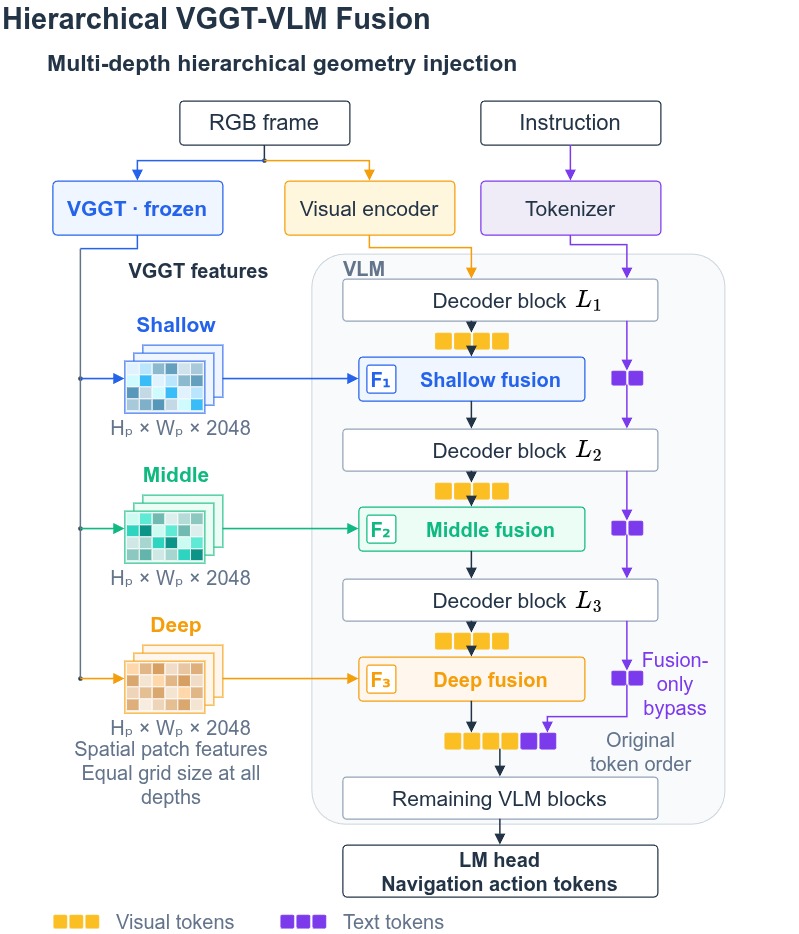}
    \caption{\textbf{Hierarchical GFM--VLM fusion across representation depth.}
    VGGT depths 11, 17, and 23 are coupled to Qwen3.5-4B decoder layers
    $L_0$, $L_1$, and $L_2$. Each branch applies RMSNorm, $2\times2$
    grouping, projection, token-wise gating, and a learned scale before
    residual addition at image-token positions.}
    \label{fig:fusion}
\end{figure}

\begin{figure*}[!t]
    \centering
    \includegraphics[width=0.85\textwidth]{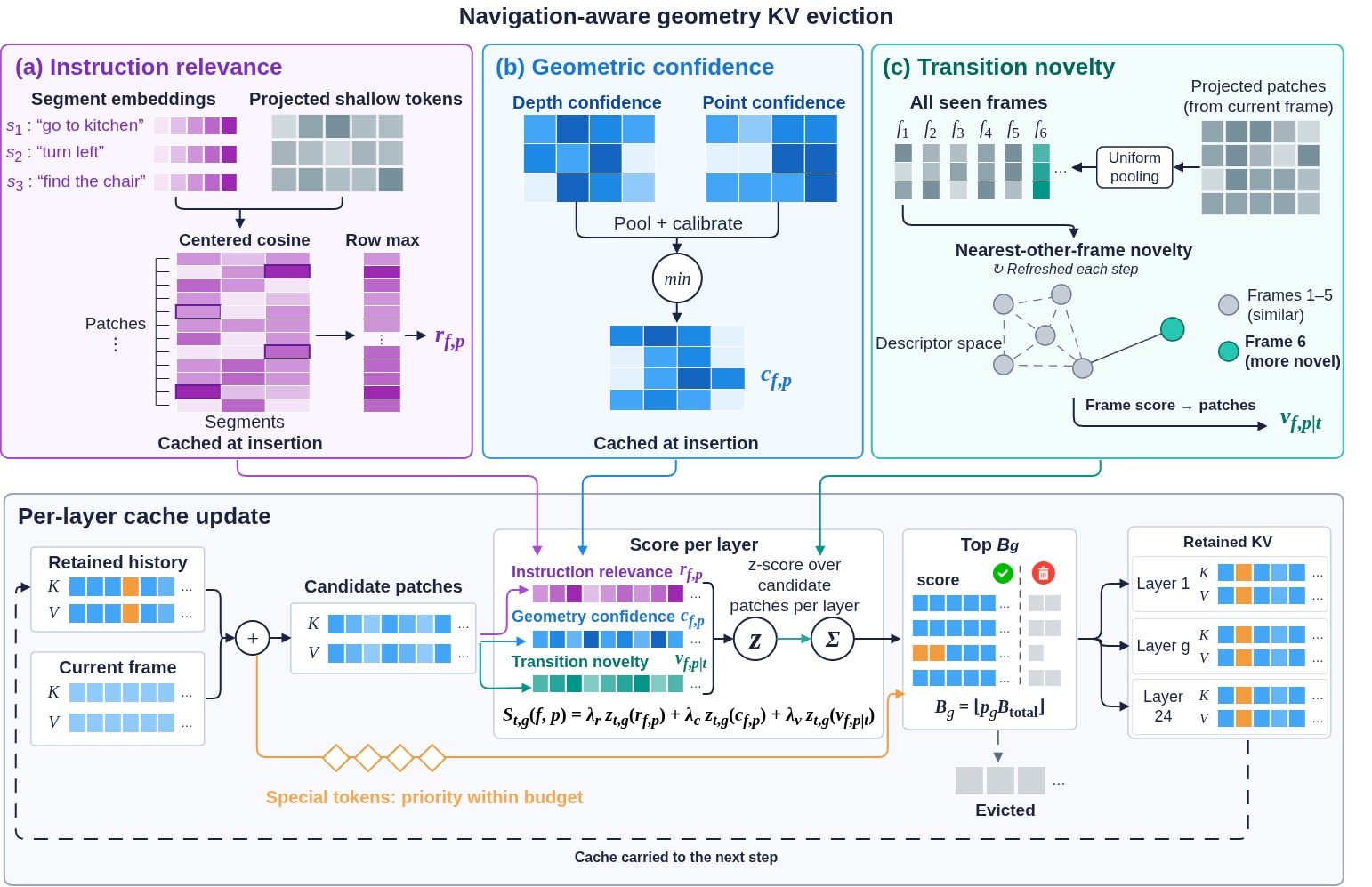}
    \caption{\textbf{Navigation-aware GFM KV retention.} Instruction relevance, geometric confidence, and transition novelty jointly determine which historical and current states remain available. Layer-specific score normalization and budgeted selection yield the geometric memory used at the next observation.}
    \label{fig:retention}
\end{figure*}

\method{} comprises two mechanisms corresponding to the depth and temporal selection problems introduced in Sec.~\ref{sec:introduction}. Hierarchical GFM--VLM fusion exposes multiple GFM depths to successive policy stages, while navigation-aware GFM memory retains historical global-attention KV states under a fixed budget. Figures~\ref{fig:fusion} and~\ref{fig:retention} illustrate these mechanisms.

\subsection{Hierarchical Geometry Fusion for Navigation Reasoning}
\label{sec:hierarchical_fusion}

Following the multi-depth interface used in SpatialStack~\cite{zhang2026spatialstack}, we extract VGGT representations at depths $m\in\{11,17,23\}$ and couple them to the first three Qwen3.5-4B decoder layers $k\in\{0,1,2\}$:
\begin{equation}
 (G_t^{11},G_t^{17},G_t^{23})
 \longrightarrow
 (L_0,L_1,L_2).
 \label{eq:depth_mapping}
\end{equation}
The selected depths span separated stages of the VGGT hierarchy and are injected early so that each geometry update can propagate through subsequent policy computation. Importantly, \method{} does not assign predefined semantic roles to these depths. Instead, the hierarchy is treated as an empirical hypothesis: latent states obtained at different stages of geometric processing may provide complementary information for navigation that is obscured when only one terminal state is repeatedly injected.

For each depth $m$, the VGGT representation is normalized and spatially aligned with the VLM image-token resolution:
\begin{equation}
 Z_t^m=\operatorname{Group}_{2\times2}\!\left(\operatorname{RMSNorm}(G_t^m)\right).
 \label{eq:group}
\end{equation}
VGGT provides a 2048-D feature at each spatial location; grouping four neighboring patches produces an 8192-D token. A depth-specific adapter maps it to the 2560-D language hidden space,
\begin{equation}
 \Phi_m:\mathbb{R}^{8192}\rightarrow\mathbb{R}^{4096}\rightarrow\mathbb{R}^{2560},
\end{equation}
while a token-wise gating network
\begin{equation}
 \Gamma_m:\mathbb{R}^{8192}\rightarrow\mathbb{R}^{2048}\rightarrow\mathbb{R}
\end{equation}
outputs a scalar gating logit for each merged token. The geometric update is
\begin{equation}
 \Delta_t^m
 =
 s_m
 \left[
 \sigma\!\left(\Gamma_m(Z_t^m)\right)
 \,\Phi_m(Z_t^m)
 \right],
 \label{eq:fusion_update}
\end{equation}
where the learned scalar $s_m$ controls the contribution of depth $m$.

Let $H_{t,\mathrm{img}}^k$ denote the image-token output of decoder layer $L_k$; geometry is injected after that layer through
\begin{equation}
 H_{t,\mathrm{img}}^{k,+}
 =
 H_{t,\mathrm{img}}^k
 +
 \Delta_t^m,
 \quad
 (m,k)\in\{(11,0),(17,1),(23,2)\}.
 \label{eq:residual}
\end{equation}
Subsequent decoder layers propagate this geometry-enriched visual context without directly overwriting text-token states.

To distinguish representation diversity from injection frequency,
Sec.~\ref{sec:depth_ablation} compares $(G^{11},G^{17},G^{23})$ with a matched control that injects $G^{23}$ at all three decoder layers.

\subsection{Navigation-Aware GFM KV Retention}
\label{sec:memory_method}
Long-horizon navigation requires geometric evidence whose utility is not determined by recency alone, yet retaining the complete history incurs memory growth with trajectory length. We therefore formulate GFM memory as \emph{navigation-conditioned token selection} under a fixed capacity. Three complementary criteria guide retention: instruction relevance, geometric confidence, and transition importance. Selection operates on the \emph{VGGT global-attention KV cache}; the Qwen/VLM cache and the CPU buffer of frame-aligned $G^{11/17/23}$ features used for hierarchical fusion (Sec.~\ref{sec:hierarchical_fusion}) are separate.

For global-attention layer $g$, the candidate memory combines retained history with the current observation:
\begin{equation}
 \mathcal{C}_t^g=M_{t-1}^g\cup(K_t^g,V_t^g).
 \label{eq:candidate_cache}
\end{equation}
Selection follows the current VGGT pass; thus, the retained states determine geometric context at $t+1$ without altering the current-frame representation.

\paragraph{Instruction-conditioned relevance.}
To associate geometry with individual navigation subgoals, we segment instructions using punctuation, sequencing markers (e.g., ``then,'' ``next,'' and ``after that''), and conjunctions introducing navigation verbs, preserving phrases such as ``next to.'' Each episode-level segment embedding $e_j$ is obtained by mean-pooling and $\lVert e_j \rVert_2 = 1$ its Qwen input-token embeddings. 

Let $u_{f,p}\in\mathbb{R}^{2560}$ denote a spatially grouped token from $G_f^{11}$ projected into Qwen space before gating and scaling. To remove shared embedding components, we center visual tokens within the frame and segment embeddings within the instruction:
\begin{equation}
    \bar{u}_f = \frac{1}{P_f}\sum_{p=1}^{P_f}u_{f,p},
    \qquad
    \bar{e} = \frac{1}{J}\sum_{j=1}^{J}e_j,
\end{equation}
where $P_f$ and $J$ denote the numbers of projected tokens and instruction segments. Relevance is defined by
\begin{equation}
    r_{f,p}
    =
    \frac{1+\max_j
    \cos\!\left(u_{f,p}-\bar{u}_f,\;e_j-\bar{e}\right)}{2}.
\end{equation}
Maximization associates each token with its most relevant segment, rather than requiring alignment with the instruction as a whole. For single-segment instructions, text-side centering is omitted to avoid a zero vector.

\paragraph{Geometric confidence.}
Instruction relevance alone does not establish geometric confidence. We therefore average-pool VGGT's depth- and point-map confidence fields onto the aligned projected-token grid and map each pooled value $\tilde{c}\in(1,\infty)$ to $1-1/\tilde{c}\in(0,1)$. Their conservative combination is
\begin{equation}
    c_{f,p} =
    \min\!\left(
        c^{\mathrm{depth}}_{f,p},
        c^{\mathrm{point}}_{f,p}
    \right).
    \label{eq:confidence}
\end{equation}
The minimum makes high confidence from both predictors necessary for a high reliability score.

\paragraph{Transition novelty.}
To reduce redundant geometric history, we quantify transition novelty: how distinct aframe's viewpoint is from every other observed frame Each frame is represented by uniformly pooled, uncentered projected shallow tokens:
\begin{equation}
    d_f = \operatorname{norm}\!\left(
        \frac{1}{P_f}\sum_{p=1}^{P_f}u_{f,p}
    \right).
    \label{eq:descriptor}
\end{equation}
Here, $\operatorname{norm}$ denotes $\ell_2$ normalization. Let $\mathcal{F}_{\leq t}$ contain all observed frames, whose descriptors remain available even after KV eviction. At each step, frame novelty is recomputed relative to its nearest other descriptor:
\begin{equation}
    \nu_{f\mid t}
    =
    \frac{
        1 -
        \max_{\substack{
            f'\in\mathcal{F}_{\leq t}\\
            f'\neq f
        }}
        \cos(d_f,d_{f'})
    }{2}.
    \label{eq:transition_novelty}
\end{equation}
We define novelty as zero when only one frame has been observed. A high value otherwise indicates that no other observed frame provides a similar geometric descriptor. All candidate patches from frame $f$ share this score:
\begin{equation}
    \nu_{f,p\mid t} = \nu_{f\mid t}.
    \label{eq:anchor}
\end{equation}

\paragraph{Joint retention score.}
Relevance and confidence are mapped to source KV patches, cached at insertion, and shared across layers; novelty is refreshed at each step. To combine these signals on comparable scales, we standardize each over the candidate patches of layer $g$, excluding special tokens:
\begin{equation}
S_{t,g}(f,p) = \lambda_r z_{t,g}(r_{f,p}) +                   \lambda_c z_{t,g}(c_{f,p})
                + \lambda_\nu z_{t,g}(\nu_{f,p|t}).
\label{eq:score}
\end{equation}
Here, $z_{t,g}$ denotes layer- and step-specific z-score normalization. Equal weights are fixed without tuning; the leave-one-signal-out ablations in \hyperref[sec:memory_exp]{Sec.~\ref*{sec:memory_exp}~(c)} assess signal contributions under the same budget, rather than weight optimality.

\paragraph{Budgeted memory selection.}
The total capacity is distributed across VGGT's 24 global-attention layers as
\begin{equation}
 B_g=\left\lfloor\rho_g B_{\mathrm{total}}\right\rfloor,
 \qquad \sum_{g=1}^{24}\rho_g=1,
 \label{eq:budget}
\end{equation}
where $\rho_g$ is derived offline from per-layer input--output cosine similarity following GHOST~\cite{chen2026ghost}. Independent selection then defines the retained memory:
\begin{equation}
 M_t^g=\operatorname{TopK}\!\left(
 \mathcal{C}_t^g,S_{t,g},B_g\right).
 \label{eq:topk_memory}
\end{equation}
Accordingly, budgets such as the ``900K'' setting reported in Sec.~\ref{sec:experiments} refer to the aggregate layer-token capacity \(B_{\mathrm{total}}\) across all 24 VGGT global-attention layers, rather than to the number of unique scene tokens.


\section{EXPERIMENTS}
\label{sec:experiments}

\subsection{Experimental Setup}
\paragraph{Benchmarks and training.}
We evaluate \method{} on the Val-Unseen splits of
R2R-CE~\cite{krantz2020beyond} and RxR-CE~\cite{ku2020room},
comprising $1,839$ and $3,669$ navigation episodes, respectively.
The navigation policy combines a Qwen3.5-4B backbone with
VGGT~\cite{wang2025vggt} as a frozen geometry foundation model.
We train jointly on the R2R and RxR training sets for approximately
100 hours using $8 \times$ NVIDIA H100 GPUs. All Qwen3.5 parameters
and geometry-fusion modules are optimized, while VGGT remains
frozen. Qwen3.5 SFT denotes the geometry-free Qwen3.5-4B policy fine-tuned on the R2R and RxR training sets, corresponding to the “None” geometry configuration in Table.~\ref{tab:geometry_ablation}. The policy receives a single RGB stream without panoramic observations, odometry, or depth input. Following the benchmark
protocols, we report success rate (SR), success weighted by path
length (SPL), navigation error (NE), oracle success (OS), and
normalized dynamic time warping (nDTW). NE is measured in meters;
all other metrics are multiplied by 100.

\paragraph{Evaluation configurations.}
We compare \method{} with state-of-the-art methods on both
benchmarks and investigate representation depth and bounded-memory
retention through ablations on R2R-CE Val-Unseen. Retention budgets
specify the total retained token capacity summed across layers,
rather than GPU memory in bytes. We distinguish the memory
occupied by the GFM KV cache from total allocated GPU memory,
reporting both separately. For the accuracy--memory comparison,
we use measured mean allocated GPU memory to characterize the
overall allocation footprint of each evaluated configuration.

\subsection{Main Benchmark Results}
\begin{table}[t]
\centering
\caption{\textbf{Comparison on R2R-CE Val-Unseen.}
Pano., Odo., Depth, and S.RGB denote panoramic observations, odometry,
depth, and a single RGB stream. External Data counts additional training
samples beyond R2R/RxR; ``--'' denotes an unreported total.
StreamVLN$^{*}$ is its oracle-navigation ablation~\cite{wei2025streamvln}.
NaVILA$^{*}$ excludes human-following data, and JanusVLN$^{*}$ uses no
additional external data~\cite{zeng2026janusvln}.
Bold denotes the best navigation metric among the listed variants.}
\label{tab:r2r_main}
\begingroup
\scriptsize
\setlength{\tabcolsep}{2.2pt}
\renewcommand{\arraystretch}{1.05}
\resizebox{\columnwidth}{!}{%
\begin{tabular}{@{}l|cccc|cccc|c@{}}
\toprule
\multirow{2}{*}{Method}
& \multicolumn{4}{c|}{Observation}
& \multicolumn{4}{c|}{R2R-CE Val-Unseen} & Training \\
\cline{2-10}
& Pano. & Odo. & Depth & S.RGB
& NE$\downarrow$ & OS$\uparrow$ & SR$\uparrow$ & SPL$\uparrow$
& External Data \\
\midrule
HPN+DN~\cite{krantz2021waypoint} & \cmark & \cmark & \cmark & & 6.31 & 40.0 & 36.0 & 34.0 & -- \\
Sim2Sim~\cite{krantz2022sim} & \cmark & \cmark & \cmark & & 6.07 & 52.0 & 43.0 & 36.0 & -- \\
VLN$\circlearrowright$BERT~\cite{hong2022bridging} & \cmark & \cmark & \cmark & & 5.74 & 53.0 & 44.0 & 39.0 & -- \\
Ego$^2$-Map~\cite{hong2023learning} & \cmark & \cmark & \cmark & & 5.54 & 56.0 & 47.0 & 41.0 & -- \\
DreamWalker~\cite{wang2023dreamwalker} & \cmark & \cmark & \cmark & & 5.53 & 59.0 & 49.0 & 44.0 & -- \\
\midrule
AO-Planner~\cite{chen2025affordances} & \cmark & & \cmark & & 5.55 & 59.0 & 47.0 & 33.0 & -- \\
g3D-LF~\cite{wang2025g3d} & & \cmark & \cmark & \cmark & 5.70 & 59.5 & 47.2 & 34.6 & -- \\
Seq2Seq~\cite{krantz2020beyond} & & & \cmark & \cmark & 7.77 & 37.0 & 25.0 & 22.0 & -- \\
NaVid-4D~\cite{liu2025vid} & & & \cmark & \cmark & 5.99 & 55.7 & 43.8 & 37.1 & -- \\
NavMorph~\cite{yao2025navmorph} & & & \cmark & \cmark & 5.75 & 56.9 & 47.9 & 33.2 & -- \\
\midrule
NaVid~\cite{zhang2024navid} & & & & \cmark & 5.47 & 49.1 & 37.4 & 35.9 & 953K \\
Sim2Real~\cite{wang2024sim} & & & & \cmark & 5.95 & 55.8 & 44.9 & 30.4 & 0K \\
StreamVLN$^{*}$~\cite{wei2025streamvln} & & & & \cmark & 5.98 & 51.3 & 45.6 & 42.3 & -- \\
Uni-NaVid~\cite{zhang2024uni} & & & & \cmark & 5.58 & 53.3 & 47.0 & 42.7 & 3577K \\
NaVILA$^{*}$~\cite{cheng2024navila} & & & & \cmark & 5.37 & 57.6 & 49.7 & 45.5 & 12574K \\
JanusVLN$^{*}$~\cite{zeng2026janusvln} & & & & \cmark & \textbf{5.17} & 58.0 & 52.8 & 49.2 & 0K \\
\rowcolor{oursrow}\textbf{\method{}(Ours)} & & & & \cmark & 5.27 & \textbf{60.7} & \textbf{55.7} & \textbf{51.4} & 0K \\
\bottomrule
\end{tabular}}
\endgroup
\end{table}

\begin{table}[t]
\centering
\caption{\textbf{Comparison on RxR-CE Val-Unseen.}
Notation follows Table~\ref{tab:r2r_main}.
StreamVLN uses its standard configuration~\cite{wei2025streamvln};
its total external sample count, including EnvDrop, is not separately
quantified. JanusVLN$^{*}$ uses no additional external data.
Bold denotes the best navigation metric among the listed variants.}
\label{tab:rxr_main}
\begingroup
\scriptsize
\setlength{\tabcolsep}{2.2pt}
\renewcommand{\arraystretch}{1.05}
\resizebox{\columnwidth}{!}{%
\begin{tabular}{@{}l|cccc|cccc|c@{}}
\toprule
\multirow{2}{*}{Method}
& \multicolumn{4}{c|}{Observation}
& \multicolumn{4}{c|}{RxR-CE Val-Unseen} & Training \\
\cline{2-10}
& Pano. & Odo. & Depth & S.RGB
& NE$\downarrow$ & SR$\uparrow$ & SPL$\uparrow$ & nDTW$\uparrow$
& External Data \\
\midrule
VLN$\circlearrowright$BERT~\cite{hong2022bridging} & \cmark & \cmark & \cmark & & 8.98 & 27.0 & 22.6 & 46.7 & -- \\
\midrule
AO-Planner~\cite{chen2025affordances} & \cmark & & \cmark & & 7.06 & 43.3 & 30.5 & 50.1 & -- \\
Seq2Seq~\cite{krantz2020beyond} & & & \cmark & \cmark & 12.10 & 13.9 & 11.9 & 30.8 & -- \\
NavMorph~\cite{yao2025navmorph} & & & \cmark & \cmark & 8.85 & 30.8 & 22.8 & 44.2 & -- \\
\midrule
Sim2Real~\cite{wang2024sim} & & & & \cmark & 8.79 & 36.7 & 25.5 & 18.1 & 0K \\
StreamVLN~\cite{wei2025streamvln} & & & & \cmark & 6.72 & 48.6 & 42.5 & 60.2 & -- \\
Uni-NaVid~\cite{zhang2024uni} & & & & \cmark & 6.24 & 48.7 & 40.9 & -- & 3577K \\
NaVILA~\cite{cheng2024navila} & & & & \cmark & 6.77 & 49.3 & 44.0 & 58.8 & 13132K \\
JanusVLN$^{*}$~\cite{zeng2026janusvln} & & & & \cmark & 6.46 & 51.4 & 44.3 & 59.1 & 0K \\
\rowcolor{oursrow}\textbf{\method{} (Ours)} & & & & \cmark & \textbf{5.64} & \textbf{54.1} & \textbf{44.7} & \textbf{61.8} & 0K \\
\bottomrule
\end{tabular}}
\endgroup
\end{table}

Tables~\ref{tab:r2r_main} and~\ref{tab:rxr_main} compare representative methods and training variants, grouped by observation requirements. On R2R-CE, \method{} achieves 55.7\% SR and 51.4\% SPL, exceeding JanusVLN$^{*}$ by 2.9 and 2.2 percentage points, respectively. OS increases from 58.0\% to 60.7\%, with comparable NE (5.27 versus 5.17\,m). On RxR-CE, \method{} achieves 54.1\% SR, 44.7\% SPL, 61.8 nDTW, and 5.64\,m NE. Relative to JanusVLN$^{*}$, these correspond to gains of 2.7 percentage points in SR, 0.4 points in SPL, and 2.7 points in nDTW, with a 0.82\,m reduction in NE.

The improvements on both benchmarks suggest that the benefits of geometry-aware navigation extend across their different instruction and trajectory distributions. These results use a single RGB stream without additional navigation training samples beyond R2R/RxR. Cross-method comparisons should nevertheless account for the sensing assumptions and external supervision reported in the tables; the following ablations isolate the contributions of representation depth and temporal retention within our model.

\subsection{Effect of Hierarchical Geometry Fusion}
\label{sec:depth_ablation}
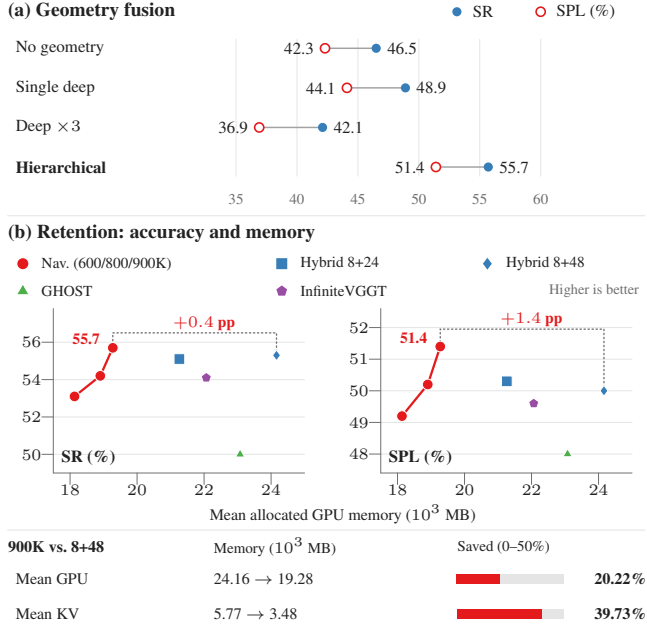
\begin{figure}[t]
\centering
\resizebox{\columnwidth}{!}{
\begingroup%
\definecolor{agSR}{HTML}{377EB8}%
\definecolor{agSPL}{HTML}{E41A1C}%
\definecolor{agMemory}{HTML}{E41A1C}%
\definecolor{agInk}{HTML}{222222}%
\definecolor{agGray}{HTML}{707070}%
\definecolor{agGrid}{HTML}{E5E5E5}%
\begin{tikzpicture}[x=1mm,y=1mm,
 every node/.style={font=\fontsize{6.3}{7}\selectfont,text=agInk,inner sep=0pt},
 line cap=round]
\path (0,0) rectangle (85.725,85.725);
\node[anchor=west,font=\bfseries\fontsize{7.3}{8}\selectfont] at (1,82.6) {(a) Geometry fusion};
\fill[agSR] (60,82.6) circle (0.6);
\node[anchor=west] at (62,82.6) {SR};
\draw[agSPL,fill=white,line width=0.6pt] (71,82.6) circle (0.6);
\node[anchor=west] at (73,82.6) {SPL (\%)};
\draw[agGrid,line width=0.25pt] (31.0,61) -- (31.0,79);
\node[font=\fontsize{5.6}{6}\selectfont,text=agGray] at (31.0,58) {35};
\draw[agGrid,line width=0.25pt] (39.0,61) -- (39.0,79);
\node[font=\fontsize{5.6}{6}\selectfont,text=agGray] at (39.0,58) {40};
\draw[agGrid,line width=0.25pt] (47.0,61) -- (47.0,79);
\node[font=\fontsize{5.6}{6}\selectfont,text=agGray] at (47.0,58) {45};
\draw[agGrid,line width=0.25pt] (55.0,61) -- (55.0,79);
\node[font=\fontsize{5.6}{6}\selectfont,text=agGray] at (55.0,58) {50};
\draw[agGrid,line width=0.25pt] (63.0,61) -- (63.0,79);
\node[font=\fontsize{5.6}{6}\selectfont,text=agGray] at (63.0,58) {55};
\draw[agGrid,line width=0.25pt] (71.0,61) -- (71.0,79);
\node[font=\fontsize{5.6}{6}\selectfont,text=agGray] at (71.0,58) {60};
\node[anchor=west] at (2,77.8) {No geometry};
\draw[agGray!65,line width=0.55pt] (42.680,77.8) -- (49.400,77.8);
\draw[agSPL,fill=white,line width=0.6pt] (42.680,77.8) circle (0.6);
\fill[agSR] (49.400,77.8) circle (0.6);
\node[anchor=east,text=agInk] at (41.280,77.8) {42.3};
\node[anchor=west,text=agInk] at (50.800,77.8) {46.5};
\node[anchor=west] at (2,72.6) {Single deep};
\draw[agGray!65,line width=0.55pt] (45.560,72.6) -- (53.240,72.6);
\draw[agSPL,fill=white,line width=0.6pt] (45.560,72.6) circle (0.6);
\fill[agSR] (53.240,72.6) circle (0.6);
\node[anchor=east,text=agInk] at (44.160,72.6) {44.1};
\node[anchor=west,text=agInk] at (54.640,72.6) {48.9};
\node[anchor=west] at (2,67.4) {Deep $\times3$};
\draw[agGray!65,line width=0.55pt] (34.040,67.4) -- (42.360,67.4);
\draw[agSPL,fill=white,line width=0.6pt] (34.040,67.4) circle (0.6);
\fill[agSR] (42.360,67.4) circle (0.6);
\node[anchor=east,text=agInk] at (32.640,67.4) {36.9};
\node[anchor=west,text=agInk] at (43.760,67.4) {42.1};
\node[anchor=west,font=\bfseries\fontsize{6.3}{7}\selectfont,text=agInk] at (2,62.2) {Hierarchical};
\draw[agGray!65,line width=0.55pt] (57.240,62.2) -- (64.120,62.2);
\draw[agSPL,fill=white,line width=0.6pt] (57.240,62.2) circle (0.6);
\fill[agSR] (64.120,62.2) circle (0.6);
\node[anchor=east,text=agInk] at (55.840,62.2) {51.4};
\node[anchor=west,text=agInk] at (65.520,62.2) {55.7};

\draw[agGrid,line width=0.4pt] (1,55.8) -- (84.7,55.8);
\node[anchor=west,font=\bfseries\fontsize{7.3}{8}\selectfont] at (1,53.5) {(b) Retention: accuracy and memory};
\definecolor{agNav}{HTML}{E41A1C}
\definecolor{agHybrid}{HTML}{377EB8}
\definecolor{agGhost}{HTML}{4DAF4A}
\definecolor{agInfinite}{HTML}{984EA3}
\draw[agNav,fill=agNav,line width=0.5pt] plot[only marks,mark=*,mark size=1.8pt] coordinates {(3,49.5)};
\node[anchor=west,font=\fontsize{5.7}{6.3}\selectfont] at (5.4,49.5) {Nav. (600/800/900K)};
\draw[agHybrid,fill=agHybrid,line width=0.5pt] plot[only marks,mark=square*,mark size=1.8pt] coordinates {(37,49.5)};
\node[anchor=west,font=\fontsize{5.7}{6.3}\selectfont] at (39.4,49.5) {Hybrid 8+24};
\draw[agHybrid,fill=agHybrid,line width=0.5pt] plot[only marks,mark=diamond*,mark size=1.8pt] coordinates {(64,49.5)};
\node[anchor=west,font=\fontsize{5.7}{6.3}\selectfont] at (66.4,49.5) {Hybrid 8+48};
\draw[agGhost,fill=agGhost,line width=0.5pt] plot[only marks,mark=triangle*,mark size=1.8pt] coordinates {(3,45.9)};
\node[anchor=west,font=\fontsize{5.7}{6.3}\selectfont] at (5.4,45.9) {GHOST};
\draw[agInfinite,fill=agInfinite,line width=0.5pt] plot[only marks,mark=pentagon*,mark size=1.8pt] coordinates {(37,45.9)};
\node[anchor=west,font=\fontsize{5.7}{6.3}\selectfont] at (39.4,45.9) {InfiniteVGGT};
\node[anchor=east,font=\fontsize{5.4}{6}\selectfont,text=agGray] at (84,45.9) {Higher is better};
\begin{axis}[
at={(7mm,22mm)},anchor=south west,scale only axis,width=33mm,height=21.6mm,
xmin=17.5,xmax=25,ymin=49,ymax=57.8,xtick={18,20,22,24},ytick={50,52,54,56},
scaled x ticks=false,
tick label style={font=\fontsize{5.8}{6.3}\selectfont,text=agInk},
axis x line*=bottom,axis y line*=left,
axis line style={agGray,line width=0.3pt},tick style={agGray},tick align=outside,
ymajorgrids=true,grid style={agGrid,line width=0.25pt},clip=true]
\draw[agGray,densely dotted,line width=0.5pt]
(axis cs:24.16194,55.3) -- (axis cs:24.16194,56.5) -- (axis cs:19.27587,56.5) -- (axis cs:19.27587,55.7);
\node[anchor=south,text=agNav,font=\bfseries\fontsize{5.8}{6.3}\selectfont,inner sep=1pt]
at (axis cs:22,56.5) {$+0.4$ pp};
\addplot[color=agNav,mark=*,mark size=2pt,line width=0.85pt,mark options={solid,fill=agNav,draw=white,line width=0.25pt}] coordinates {(18.13358,53.1) (18.90307,54.2) (19.27587,55.7)};
\addplot[color=agHybrid,mark=square*,mark size=2pt,only marks,mark options={solid,fill=agHybrid,draw=white,line width=0.25pt}] coordinates {(21.26314,55.1)};
\addplot[color=agHybrid,mark=diamond*,mark size=2pt,only marks,mark options={solid,fill=agHybrid,draw=white,line width=0.25pt}] coordinates {(24.16194,55.3)};
\addplot[color=agGhost,mark=triangle*,mark size=2pt,only marks,mark options={solid,fill=agGhost,draw=white,line width=0.25pt}] coordinates {(23.0755,50.0)};
\addplot[color=agInfinite,mark=pentagon*,mark size=2pt,only marks,mark options={solid,fill=agInfinite,draw=white,line width=0.25pt}] coordinates {(22.06632,54.1)};
\node[anchor=south east,text=agNav,font=\bfseries\fontsize{5.8}{6.3}\selectfont,inner sep=1.2pt]
at (axis cs:19.0,55.7) {55.7};
\end{axis}
\node[anchor=south west,font=\bfseries\fontsize{6.3}{7}\selectfont] at (8,23) {SR (\%)};
\begin{axis}[
at={(50mm,22mm)},anchor=south west,scale only axis,width=33mm,height=21.6mm,
xmin=17.5,xmax=25,ymin=47.4,ymax=52.6,xtick={18,20,22,24},ytick={48,49,50,51,52},
scaled x ticks=false,
tick label style={font=\fontsize{5.8}{6.3}\selectfont,text=agInk},
axis x line*=bottom,axis y line*=left,
axis line style={agGray,line width=0.3pt},tick style={agGray},tick align=outside,
ymajorgrids=true,grid style={agGrid,line width=0.25pt},clip=true]
\draw[agGray,densely dotted,line width=0.5pt]
(axis cs:24.16194,50.0) -- (axis cs:24.16194,51.95) -- (axis cs:19.27587,51.95) -- (axis cs:19.27587,51.4);
\node[anchor=south,text=agNav,font=\bfseries\fontsize{5.8}{6.3}\selectfont,inner sep=1pt]
at (axis cs:22,51.95) {$+1.4$ pp};
\addplot[color=agNav,mark=*,mark size=2pt,line width=0.85pt,mark options={solid,fill=agNav,draw=white,line width=0.25pt}] coordinates {(18.13358,49.2) (18.90307,50.2) (19.27587,51.4)};
\addplot[color=agHybrid,mark=square*,mark size=2pt,only marks,mark options={solid,fill=agHybrid,draw=white,line width=0.25pt}] coordinates {(21.26314,50.3)};
\addplot[color=agHybrid,mark=diamond*,mark size=2pt,only marks,mark options={solid,fill=agHybrid,draw=white,line width=0.25pt}] coordinates {(24.16194,50.0)};
\addplot[color=agGhost,mark=triangle*,mark size=2pt,only marks,mark options={solid,fill=agGhost,draw=white,line width=0.25pt}] coordinates {(23.0755,48.0)};
\addplot[color=agInfinite,mark=pentagon*,mark size=2pt,only marks,mark options={solid,fill=agInfinite,draw=white,line width=0.25pt}] coordinates {(22.06632,49.6)};
\node[anchor=south east,text=agNav,font=\bfseries\fontsize{5.8}{6.3}\selectfont,inner sep=1.2pt]
at (axis cs:19.0,51.4) {51.4};
\end{axis}
\node[anchor=south west,font=\bfseries\fontsize{6.3}{7}\selectfont] at (51,23) {SPL (\%)};
\node[font=\fontsize{6}{6.5}\selectfont] at (45,16.6) {Mean allocated GPU memory ($10^3$ MB)};
\draw[agGrid,line width=0.4pt] (1,14.5) -- (84.7,14.5);
\node[anchor=west,font=\bfseries\fontsize{6}{6.5}\selectfont] at (1,12.3) {900K vs. 8+48};
\node[anchor=west,font=\fontsize{5.6}{6}\selectfont] at (28,12.3) {Memory ($10^3$ MB)};
\node[anchor=west,font=\fontsize{5.6}{6}\selectfont] at (60,12.3) {Saved (0--50\%)};
\node[anchor=west,font=\fontsize{6}{6.5}\selectfont] at (2,8.1) {Mean GPU};
\node[anchor=west,font=\fontsize{6}{6.5}\selectfont] at (28,8.1) {24.16 $\to$ 19.28};
\fill[agGrid] (60,7.45) rectangle (74,8.75);
\fill[agMemory] (60,7.45) rectangle (65.6622,8.75);
\node[anchor=east,text=agInk,font=\bfseries\fontsize{6}{6.5}\selectfont] at (85,8.1) {20.22\%};
\node[anchor=west,font=\fontsize{6}{6.5}\selectfont] at (2,3.5) {Mean KV};
\node[anchor=west,font=\fontsize{6}{6.5}\selectfont] at (28,3.5) {5.77 $\to$ 3.48};
\fill[agGrid] (60,2.85) rectangle (74,4.15);
\fill[agMemory] (60,2.85) rectangle (71.1253,4.15);
\node[anchor=east,text=agInk,font=\bfseries\fontsize{6}{6.5}\selectfont] at (85,3.5) {39.73\%};
\pgfresetboundingbox
\path[use as bounding box] (0,0) rectangle (85.725,85.725);
\end{tikzpicture}%
\endgroup%
}
\caption{\textbf{Geometry fusion and retention on R2R-CE Val-Unseen.}
\textbf{(a)} SR/SPL for geometry-fusion ablations.
\textbf{(b)} SR/SPL versus mean allocated GPU memory; the legend identifies methods.
Red lines connect 600K, 800K, and 900K aggregate layer-token budgets.
Dotted brackets mark 900K gains over Hybrid Inc.\ (8+48), in percentage points.
The bottom strip reports corresponding GPU and GFM-KV memory savings relative to that baseline.}
\label{fig:results_summary}
\end{figure}

We test whether access to multiple geometric representation depths improves navigation beyond increasing fusion frequency. Table~\ref{tab:geometry_ablation} and Fig.~\ref{fig:results_summary}(a) compare hierarchical fusion with a geometry-free policy, a single terminal-feature injection, and repeated injection of the terminal representation at the same three fusion sites. Hierarchical fusion improves SR/SPL by 9.2/9.1 percentage points over the geometry-free policy and 6.8/7.3 points over single-deep fusion. The largest gains, 13.6/14.5 points, occur over Deep$\times3$, which matches the number and positions of fusion operations. Thus, repeated terminal-feature injection does not recover the benefit of multi-depth access. These results support selecting geometric representations across depth, without requiring fixed semantic interpretations of individual layers. The improvement is strongest in success-based metrics; the geometry-free policy retains a slightly lower NE (5.19 versus 5.27\,m). 

\begin{table}[t]
\centering
\caption{\textbf{Geometry fusion on R2R-CE Val-Unseen.}
Deep$\times3$ matches the three fusion locations of hierarchical fusion
while reusing the terminal representation.}
\label{tab:geometry_ablation}
\scriptsize
\setlength{\tabcolsep}{3.3pt}
\begin{tabular}{lcccc}
\toprule
Geometry & SR$\uparrow$ & SPL$\uparrow$ & OS$\uparrow$ & NE$\downarrow$\\
\midrule
None & 46.5 & 42.3 & 56.0 & \textbf{5.19}\\
Single deep $G^{23}{\rightarrow}L_2$ & 48.9 & 44.1 & 56.0 & 5.28\\
Deep$\times3$ $G^{23}$ & 42.1 & 36.9 & 56.7 & 6.64\\
\rowcolor{oursrow}\textbf{Hier.} $G^{11/17/23}$ & \textbf{55.7} & \textbf{51.4} & \textbf{60.7} & 5.27\\
\bottomrule
\end{tabular}
\end{table}

\definecolor{agMemoryRow}{HTML}{EDF4FA}
\subsection{Navigation-Aware Geometric Memory}
\label{sec:memory_exp}
We evaluate selected retention strategies, vary the retained-token
budget, and ablate the navigation-aware scoring components.

\paragraph{Comparison of retention strategies.}
\label{sec:matched_memory}
Navigation-aware retention improves navigation success and path
efficiency with lower memory usage than the evaluated temporal
baselines (Table~\ref{tab:matched_memory}). At a similar measured KV
footprint, Nav.\ 900K surpasses Hybrid Inc.\ (8+24) by 0.6 percentage
points in SR and 1.1 points in SPL, using 3.33\% less mean GFM-KV
memory and 9.35\% less mean allocated GPU memory. Relative to
Hybrid Inc.\ (8+48), it improves SR/SPL by 0.4/1.4 points while
reducing KV and allocated GPU memory by 39.73\% and 20.22\%,
respectively. These comparisons support navigation-aware selection
as an effective approach to improving SR/SPL with a smaller
geometric memory footprint. Nav.\ 900K achieves the highest SR
(55.7\%) and SPL (51.4\%) among the selected configurations,
while temporal baselines retain the best OS and NE.

\paragraph{Retention-budget sensitivity.}
Navigation performance improves consistently with retention capacity
(Figure~\ref{fig:results_summary}(b)). Increasing the budget from
600K to 800K and 900K raises SR from 53.1\% to 54.2\% and 55.7\%,
and SPL from 49.2\% to 50.2\% and 51.4\%, respectively.
Moving from 600K to 900K yields gains of 2.6 SR points and
2.2 SPL points for an additional 1142.29 MB of mean allocated
GPU memory. The budget therefore offers an accuracy--memory
trade-off, with 900K achieving the strongest SR/SPL among
the tested capacities.

\begin{table}[t]
\centering
\caption{\textbf{Selected retention strategies on R2R-CE Val-Unseen.}
Nav. denotes our navigation-aware retention. KV and Alloc. report mean
GFM-KV and allocated GPU memory in MB. Bold marks the best displayed value in each column. Hybrid Incr. denotes hybrid incremental update.
The notation $8+24$ and $8+48$ denotes retaining the first 8 observations
together with the most recent 24 or 48 observations, respectively.
GHOST and InfiniteVGGT use fixed budgets of \textbf{1m2} and
\textbf{1m2}, respectively.}
\label{tab:matched_memory}
\begingroup
\scriptsize
\setlength{\tabcolsep}{2pt}
\renewcommand{\arraystretch}{1.08}
\resizebox{\columnwidth}{!}{%
\begin{tabular}{@{}lcccccc@{}}
\toprule
Retention & SR$\uparrow$ & SPL$\uparrow$ & OS$\uparrow$ & NE$\downarrow$ & KV$\downarrow$ & Alloc.$\downarrow$\\
\midrule
Hybrid Inc. (8+24)~\cite{zeng2026janusvln} & 55.1 & 50.3 & \textbf{62.4} & 5.30 & 3595.53 & 21263.14 \\
Hybrid Inc. (8+48)~\cite{zeng2026janusvln} & 55.3 & 50.0 & 61.5 & \textbf{5.21} & 5767.55 & 24161.94 \\
GHOST~\cite{chen2026ghost} & 50.0 & 48.0 & 54.9 & 5.55 & 6758.17 & 23075.50 \\
InfiniteVGGT~\cite{yuan2026infinitevggt} & 54.1 & 49.6 & 60.3 & 5.30 & 6854.04 & 22066.32 \\
\midrule
Nav. 600K & 53.1 & 49.2 & 59.2 & 5.42 & \textbf{2340.81} & \textbf{18133.58} \\
Nav. 800K & 54.2 & 50.2 & 59.5 & 5.38 & 3105.28 & 18903.07 \\
\rowcolor{agMemoryRow}
Nav. 900K & \textbf{55.7} & \textbf{51.4} & 60.7 & 5.27 & 3475.92 & 19275.87 \\
\bottomrule
\end{tabular}}
\endgroup
\end{table}

\paragraph{Contribution of the retention signals.}
\label{sec:signal_ablation}
Each component of the navigation-aware score contributes to
retention quality at a fixed memory budget
(Table~\ref{tab:signal_ablation}). Removing instruction relevance,
geometric confidence, or transition importance reduces both SR
and SPL, while mean GFM-KV memory varies by only 0.20 MB.
Removing confidence causes the largest SR decrease (2.1 percentage
points), followed by instruction relevance (1.5 points) and
transition importance (1.4 points). The complete score therefore
outperforms every tested two-signal variant at approximately
constant KV memory, supporting the proposed combination of
instruction and geometric cues for selecting useful navigation
history.
\begin{table}[t]
\centering
\caption{\textbf{Leave-one-signal-out ablation at 900K on R2R-CE Val-Unseen.}
All signals denotes the complete retention score. KV reports mean
GFM-KV memory in MB. Bold marks the best navigation metric in each column.}
\label{tab:signal_ablation}
\begingroup
\scriptsize
\setlength{\tabcolsep}{3.2pt}
\begin{tabular}{@{}lccccc@{}}
\toprule
Retention score & SR$\uparrow$ & SPL$\uparrow$ & OS$\uparrow$ & NE$\downarrow$ & KV (MB)\\
\midrule
\rowcolor{agMemoryRow}
All signals & \textbf{55.7} & \textbf{51.4} & \textbf{60.7} & \textbf{5.27} & 3475.92\\
w/o Transition & 54.3 & 50.3 & 60.1 & 5.41 & 3475.96\\
w/o Confidence & 53.6 & 49.6 & 59.4 & 5.48 & 3475.85\\
w/o Instruction & 54.2 & 50.1 & 59.2 & 5.45 & 3475.76\\
\bottomrule
\end{tabular}
\endgroup
\end{table}
\subsection{Qualitative Analysis and Real-World Deployment}
\begin{figure*}[t]
\centering
\includegraphics[width=0.8\textwidth]{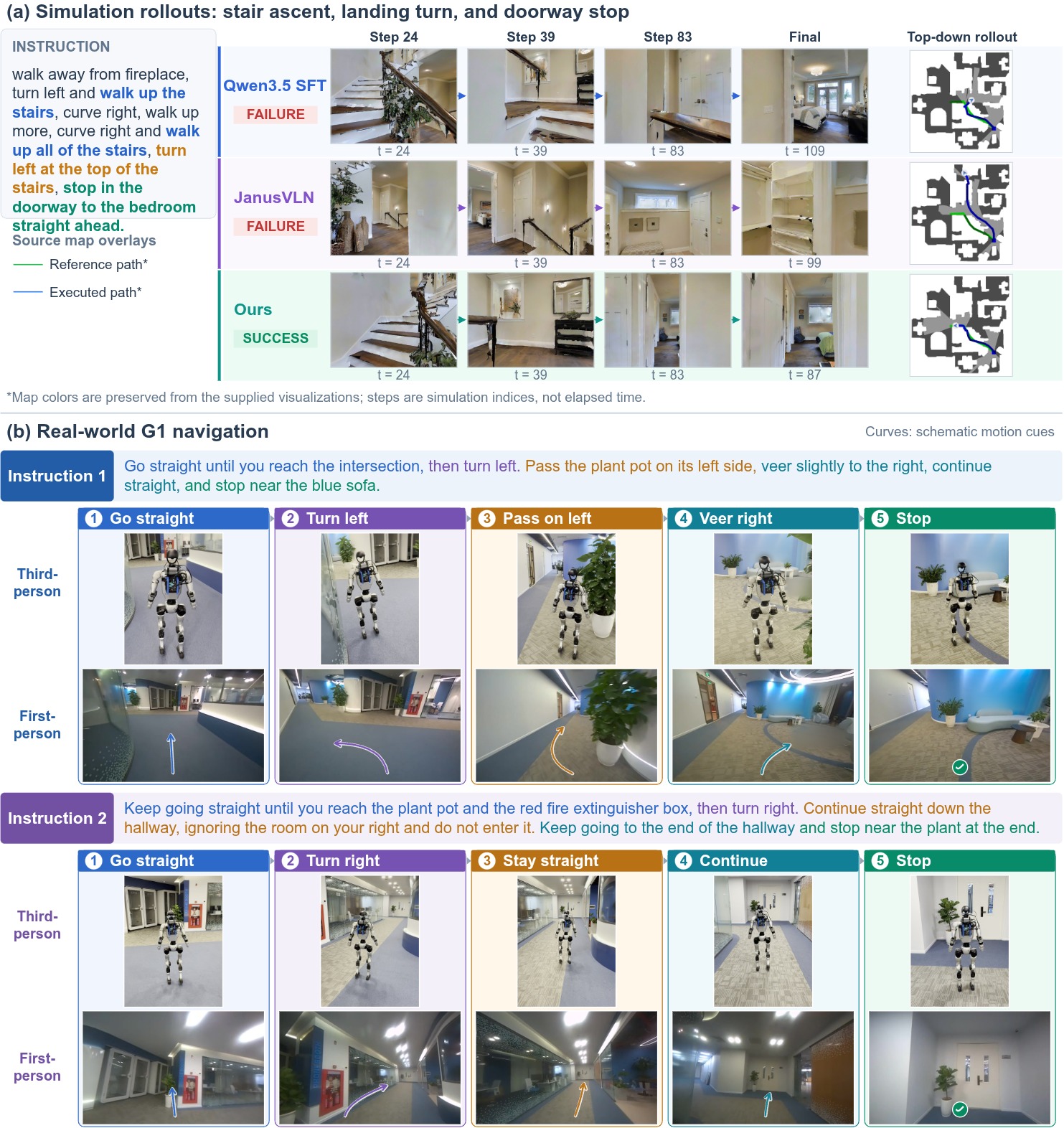}
\caption{\textbf{Simulation rollouts and humanoid deployment.}
Top: an R2R-CE Val-Unseen route requiring staircase traversal, multiple
direction changes, and precise stopping. \method{} reaches the target,
whereas Qwen3.5 SFT and JanusVLN fail to complete this route.
Bottom: representative indoor navigation sequences on a Unitree G1
equipped with a ZED X Mini camera.}
\label{fig:rollout_g1}
\end{figure*}

Figure~\ref{fig:rollout_g1} shows simulation rollouts and physical
deployment. The selected R2R-CE episode involves walking away from a fireplace, climbing a curved staircase, and stopping at a bedroom
doorway. \method{} reaches the target, whereas Qwen3.5 SFT and JanusVLN
fail, illustrating multi-step instruction following alongside the
controlled ablations. We also deploy \method{} on a Unitree G1 humanoid
with a ZED X Mini camera. Jetson AGX Orin preprocesses images
and communicates with a remote NVIDIA RTX A6000 workstation for policy
inference. The policy uses a single RGB stream and the same geometric
reasoning framework as in simulation. Two indoor trials involve
landmark-conditioned turns, passing a plant on the instructed side,
bypassing an irrelevant doorway, and stopping near designated targets.
Paired third-person and first-person views align the robot's progress
with instruction segments, demonstrating physical execution beyond
simulation.

\section{CONCLUSION}
We presented \method{}, a streaming VLN framework combining multi-depth GFM--VLM fusion with navigation-aware bounded GFM-KV retention. Evaluations on R2R-CE and RxR-CE demonstrate strong navigation performance, with R2R-CE ablations favoring multi-depth representations over single-terminal and repeated-terminal fusion. Navigation-aware retention improves SR/SPL over temporal selection at a similar measured KV footprint and substantially reduces geometric memory relative to larger-memory temporal retention. Deployment on a Unitree G1 demonstrates physical execution, supporting the use of selective geometric representations and historical evidence for embodied navigation. A limitation is the use of equal retention coefficients without optimizing their relative weights. The signal ablations do not establish optimal weighting; tuning or learning these coefficients remains future work.


\balance
\bibliographystyle{ieeetr}
\bibliography{references}
\end{document}